\documentclass[conference]{IEEEtran}
\IEEEoverridecommandlockouts

\usepackage{algorithmic}
\usepackage{amsmath,amssymb,amsfonts}
\usepackage{booktabs}
\usepackage{cite}
\usepackage{colortbl}
\usepackage{graphicx}
\usepackage{subcaption}
\usepackage{textcomp}
\usepackage{url}
\usepackage{wrapfig}
\usepackage{xcolor}
\usepackage{xspace}

\usepackage[breaklinks]{hyperref}
\usepackage{xurl}
\hypersetup{
    colorlinks=false,
    linkcolor=black,
    filecolor=black,      
    urlcolor=black,
}

\def\BibTeX{{\rm B\kern-.05em{\sc i\kern-.025em b}\kern-.08em
    T\kern-.1667em\lower.7ex\hbox{E}\kern-.125emX}}

\newcommand{\project}{ExposureEngine\xspace}

\usepackage[acronym]{glossaries}
\newacronym{obb}{OBB}{Oriented Bounding Box}
\newacronym{hbb}{HBB}{Horizontal Bounding Box}
\newacronym{ood}{OOD}{Oriented Object Detection}
\newacronym{llm}{LLM}{Large Language Model}
\newacronym{sef}{SEF}{Swedish Elite Football}
\newacronym{hls}{HLS}{HTTP Live Streaming}
\newacronym{fl}{FL}{Focal Loss}
\newacronym{bce}{BCE}{Binary Cross Entropy}
\newacronym{vfl}{VFL}{Varifocal Loss}
\newacronym{tr}{TR}{Tightness Ratio}
\newacronym{on}{ON}{Orientation Necessity}

\begin{document}
\bstctlcite{IEEEexample:BSTcontrol}

\title{\project: Oriented Logo Detection and Sponsor Visibility Analytics in Sports Broadcasts}

\newcommand{\simulamet}{SimulaMet, Norway}
\newcommand{\forzasys}{Forzasys, Norway}
\newcommand{\oslomet}{Oslo Metropolitan University, Norway}
\newcommand{\uit}{UiT, Norway}

\author{
    \IEEEauthorblockN{
        Mehdi Houshmand Sarkhoosh\IEEEauthorrefmark{3}\IEEEauthorrefmark{4},
        Fr{\o}y {\O}ye\IEEEauthorrefmark{3},
        Henrik Nestor S{\o}rlie\IEEEauthorrefmark{3},
        Nam Hoang Vu\IEEEauthorrefmark{3}, \\ 
        Dag Johansen\IEEEauthorrefmark{5},
        Cise Midoglu\IEEEauthorrefmark{4},
        Tomas Kupka\IEEEauthorrefmark{4}, and
        P{\aa}l Halvorsen\IEEEauthorrefmark{1}\IEEEauthorrefmark{3}\IEEEauthorrefmark{4}
    }\vspace{1mm}
    \IEEEauthorblockA{
        \IEEEauthorrefmark{4}\forzasys \ \ \ \ \
        \IEEEauthorrefmark{1}\simulamet \ \ \ \ \
        \IEEEauthorrefmark{3}\oslomet  \ \ \ \ \
        \IEEEauthorrefmark{5}\uit
    }
}

\maketitle

\begin{abstract} 
Quantifying sponsor visibility in sports broadcasts is a critical marketing task traditionally hindered by manual, subjective, and unscalable analysis methods. While automated systems offer an alternative, their reliance on axis-aligned \gls{hbb} leads to inaccurate exposure metrics when logos appear rotated or skewed due to dynamic camera angles and perspective distortions. This paper introduces \project, an end-to-end system designed for accurate, rotation-aware sponsor visibility analytics in sports broadcasts, demonstrated in a soccer case study. Our approach predicts  \gls{obb} to provide a geometrically precise fit to each logo regardless of the orientation on-screen. To train and evaluate our detector, we developed a new dataset comprising 1,103 frames from Swedish elite soccer, featuring 670 unique sponsor logos annotated with \gls{obb}s. Our model achieves a mean Average Precision (mAP@0.5) of 0.859, with a precision of 0.96 and recall of 0.87, demonstrating robust performance in localizing logos under diverse broadcast conditions. The system integrates these detections into an analytical pipeline that calculates precise visibility metrics, such as exposure duration and on-screen coverage. Furthermore, we incorporate a language-driven agentic layer, enabling users to generate reports, summaries, and media content through natural language queries. The complete system, including the dataset and the analytics dashboard, provides a comprehensive solution for auditable and interpretable sponsor measurement in sports media. An overview of the \project\ is available online.\footnote{\url{https://youtu.be/tRw6OBISuW4}}
\end{abstract}

\begin{IEEEkeywords}
Oriented Object Detection, Sponsorship Analytics, Logo Detection, Sports Broadcasting, Deep Learning
\end{IEEEkeywords}

\glsresetall

\section{Introduction}

Sponsor visibility is a key metric in professional sports. Rights holders and broadcasters assess how often and how clearly sponsor logos appear during a match to justify advertising rates and quantify return on investment~\cite{shi2024sports}. Traditionally, this task was done manually by annotators reviewing footage and logging logo appearances. This process is time-consuming, post hoc, subjective, and difficult to scale.

\begin{figure}[t]
    \centering
    \includegraphics[width=1\columnwidth, trim=10 60 10 0, clip]{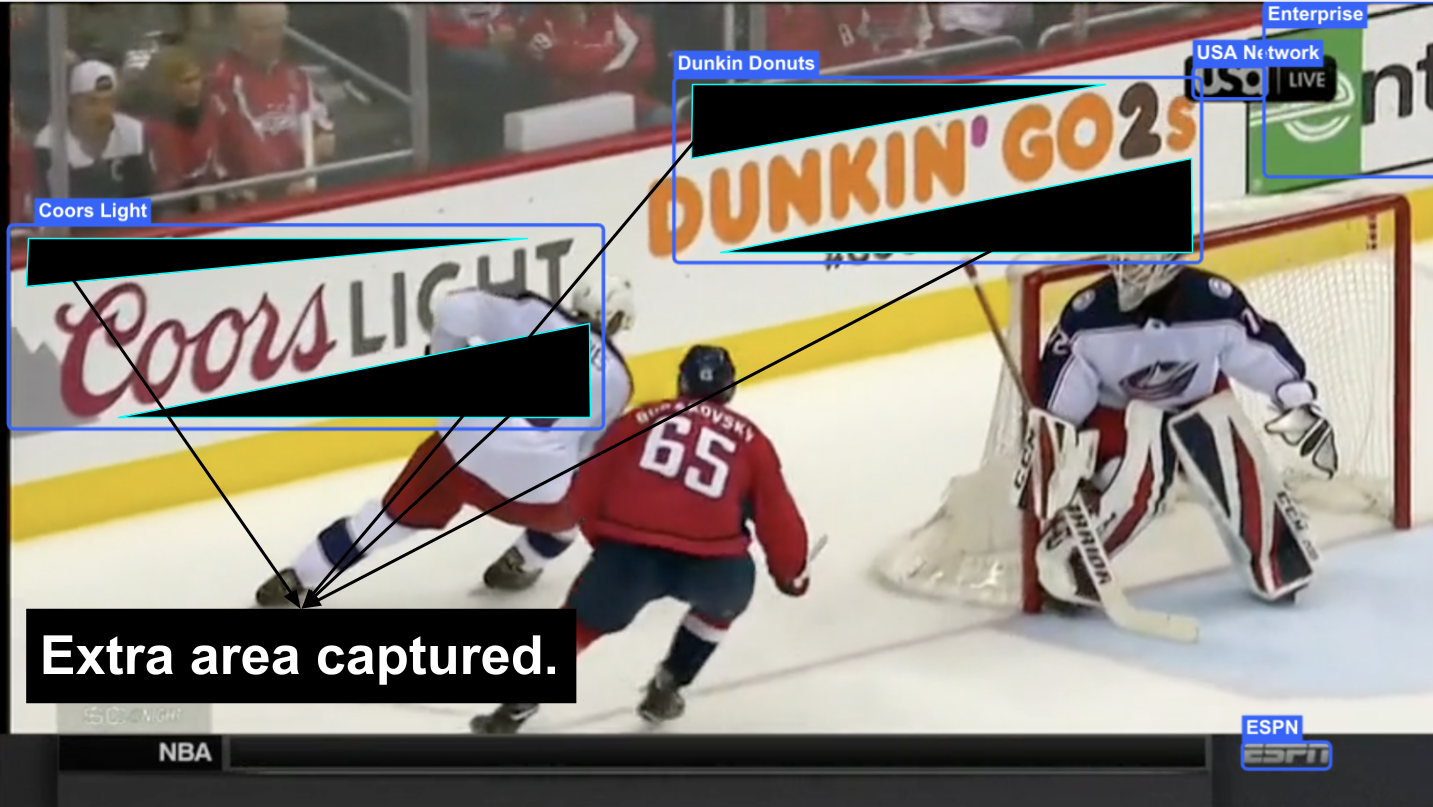}
    \caption{Example from a commercial logo detection solution using HBB~\cite{thehive_ai}. The HBB captures background regions around the rotated logo, leading to inaccurate size estimation.}
    \label{fig:obb_motivation}
\end{figure}

Computer vision offers a scalable alternative by detecting logos frame-by-frame in broadcast video~\cite{gudauskas2024providing}. Early systems relied on object detectors using  a \gls{hbb} approach. These models detect logos, but often misrepresent their shape and position when the logos appear rotated or skewed, common in broadcast footage due to dynamic camera angles, jersey folds, or perspective distortions. Axis-aligned boxes tend to cover both the logo and surrounding background, leading to overestimation of logo size and inaccurate visibility metrics such as screen prominence.
This misalignment not only impacts exposure analytics but also reduces precision in downstream tasks like logo removal, masking, or editing. Figure~\ref{fig:obb_motivation} shows an example where a \gls{hbb} includes non-logo pixels due to lack of rotation handling.

To address these issues, we present the \textit{\project} system for oriented logo detection and sponsor visibility analytics in sports broadcasts, using soccer as a case study. Our method uses a object detection model trained to predict \gls{obb}, i.e., a bounding box without the extra, wasted black areas in Figure~\ref{fig:obb_motivation}. 
We used a custom dataset of 1,103 annotated frames from the 2024 Swedish elite soccer league. The dataset covers 670 unique sponsor logos across a variety of stadiums, teams, and camera conditions. 
The \gls{obb}-based detector produces rotation-aware outputs that align more precisely with the logo shapes, enabling accurate measurement of on-screen position, size, and exposure duration. All visibility metrics are derived directly from model predictions without requiring manual correction or post-processing. 
To support analytics and human interaction, the system includes a language-based layer powered by a \gls{llm}. This module allows users to issue natural language queries and generate summaries or reports from the detection data~\cite{Sanguinetti_Gonzalez_2025}. It supports operations such as ranking logos by exposure, retrieving sponsor-specific statistics, preparing shareable media content, and generating concise video summaries of sponsor visibility.

This work makes four main contributions:  
(i) an \gls{obb} logo detection model based on YOLOv11;  
(ii) a public dataset of diverse sponsor logos from soccer footage, available\footnote{\url{https://huggingface.co/datasets/SimulaMet-HOST/ExposureEngine}};  
(iii) An analytics dashboard for sponsorship; and  
(iv) a multi-layer agent system built on the LangGraph~\cite{langgraph} agent framework that interprets user queries and generates appropriate actions based on the results.

\section{Background}

Automatic logo detection has become a key task in sports media analytics. Early approaches relied on manual annotation, where analysts logged logo appearances across match footage~\cite{shi2024sports}. This process is slow, expensive, and prone to inconsistency. With the rise of deep learning, object detection models have been increasingly used to automate this task~\cite{gudauskas2024providing,acm_logo_survey}.

Commercial systems like Relo Metrics~\cite{RELO}, Blinkfire~\cite{blinkfire}, and Hive~\cite{thehive_ai} use CNN-based detectors such as YOLO and Faster R-CNN to track logos in real time. One-stage detectors like YOLO are often preferred due to their speed and competitive accuracy~\cite{deepsense_ai_logo}. These models can support sponsor dashboards that measure exposure metrics across entire matches~\cite{acm_logo_survey}, offering scalable alternatives to manual tagging.

A critical limitation in most existing systems is their use of axis-aligned \gls{hbb}. These rectangular boxes assume upright objects with no rotation, which is rarely the case in practice. Logos in sports footage often appear tilted or warped due to camera perspectives, curved jerseys, or slanted advertising surfaces. When using \gls{hbb}s, detections frequently include extra background pixels that do not belong to the logo. This misalignment leads to overestimated exposure metrics and imprecise localization (see Figure~\ref{fig:obb_motivation}). In downstream tasks such as logo removal or inpainting, these inaccuracies introduce visual artifacts and degrade editing quality~\cite{deep_erase}.

In addition to these modeling limitations, the datasets available for training and evaluation are often not well aligned with the demands of broadcast footage. Logos in live sports appear at a range of scales and under conditions like motion blur, occlusion, and deformation. While public datasets such as LogoDet-3K~\cite{LogoDet-3K} and Flickr Sports Logo~\cite{flicker} offer large collections of logo instances, they are not tailored to the visual characteristics of televised matches. More importantly, these datasets primarily use \gls{hbb} annotations and rarely contain logos in their rotated or distorted forms. Existing commercial datasets may address these limitations, but they are proprietary and not available for academic benchmarking. At present, there is no open dataset that provides \gls{obb} annotations for sponsor logos appearing in real soccer broadcasts.

To better handle orientation variation, recent work in computer vision has introduced \gls{ood} models that estimate not only position and size, but also rotation angle~\cite{oriented_ood_review, OOD_2016, OOD_2023}. \gls{ood} methods have proven effective in domains where objects appear at arbitrary angles, such as aerial surveillance and scene text detection~\cite{OOB_survey}. Architectures like YOLOv8 and YOLOv11 extend their output heads to include angle regression, enabling real-time, rotation-aware inference~\cite{henrynavarro2024yolo}. Evaluations on benchmarks like DOTA~\cite{DOTA} and HRSC2016~\cite{HRSC2016} show clear performance gains over \gls{hbb}-based methods. Despite these advances, their application to sports broadcasts remains underexplored, due in part to the lack of annotated data in this domain.

Another layer of complexity in sponsor tracking involves how detection results are translated into useful insights. Metrics such as screen time, logo prominence, and spatial distribution must be aggregated and presented in an interpretable format. Traditional systems rely on static dashboards or custom scripts~\cite{deepsense_ai_logo}. More recently, \gls{llm}s have been introduced as a way to make analytics more accessible. By using \gls{llm}s to interpret structured outputs, systems can support human queries like “Which sponsor was most visible in the second half?” or “Summarize Adidas exposure across both halves”~\cite{arxiv_visiongpt}. These interfaces offer a more natural way to explore data, especially for non-technical stakeholders.

While prior work has addressed logo detection, orientation modeling, and analytics independently, few systems combine these components in a unified solution. The literature still lacks an open \gls{obb}-annotated dataset specific to soccer broadcasts, as well as a full-stack pipeline that supports both rotation-aware detection and language-guided sponsor analysis. This motivates the development of a system that brings together oriented detection, large-scale class handling, and human-centered analytics in the context of sports video.

\section{Dataset}\label{sec:dataset}

To support the development and evaluation of rotation-aware sponsor logo detection systems, we constructed a dedicated dataset based on 97 professional soccer highlight clips from the 2024 Swedish men's elite league season. Videos were selected to cover a range of game events and corresponding camera perspectives, e.g., close-ups for red cards and wide shots for goal celebrations, to ensure varied visual contexts. From these, representative frames were sampled at 1~FPS, filtered to remove near-duplicates, and manually reviewed to retain only those with visible sponsor logos. This process resulted in a curated set of 1,103 frames, each named using a consistent scheme encoding the match and frame index. Table~\ref{tab:dataset_composition} summarizes the dataset composition.

\begin{table}[ht]
    \centering
    \caption{Overview of the dataset composition.}
    \label{tab:dataset_composition}
    \begin{tabular}{ll}
        \toprule
        Property & Value \\
        \midrule
        Number of matches & 32 \\
        Number of teams & 16 \\
        Event types & Goal, Shot, Yellow card, Offside, Substitution \\
        Videos per match & 3 (typically one per event type) \\
        Total video clips & 97 \\
        Season & 2024 \\
    \bottomrule
    \end{tabular}
\end{table}

\begin{figure}[t]
    \centering
    \includegraphics[width=0.92\columnwidth, trim=0 0 0 0, clip]{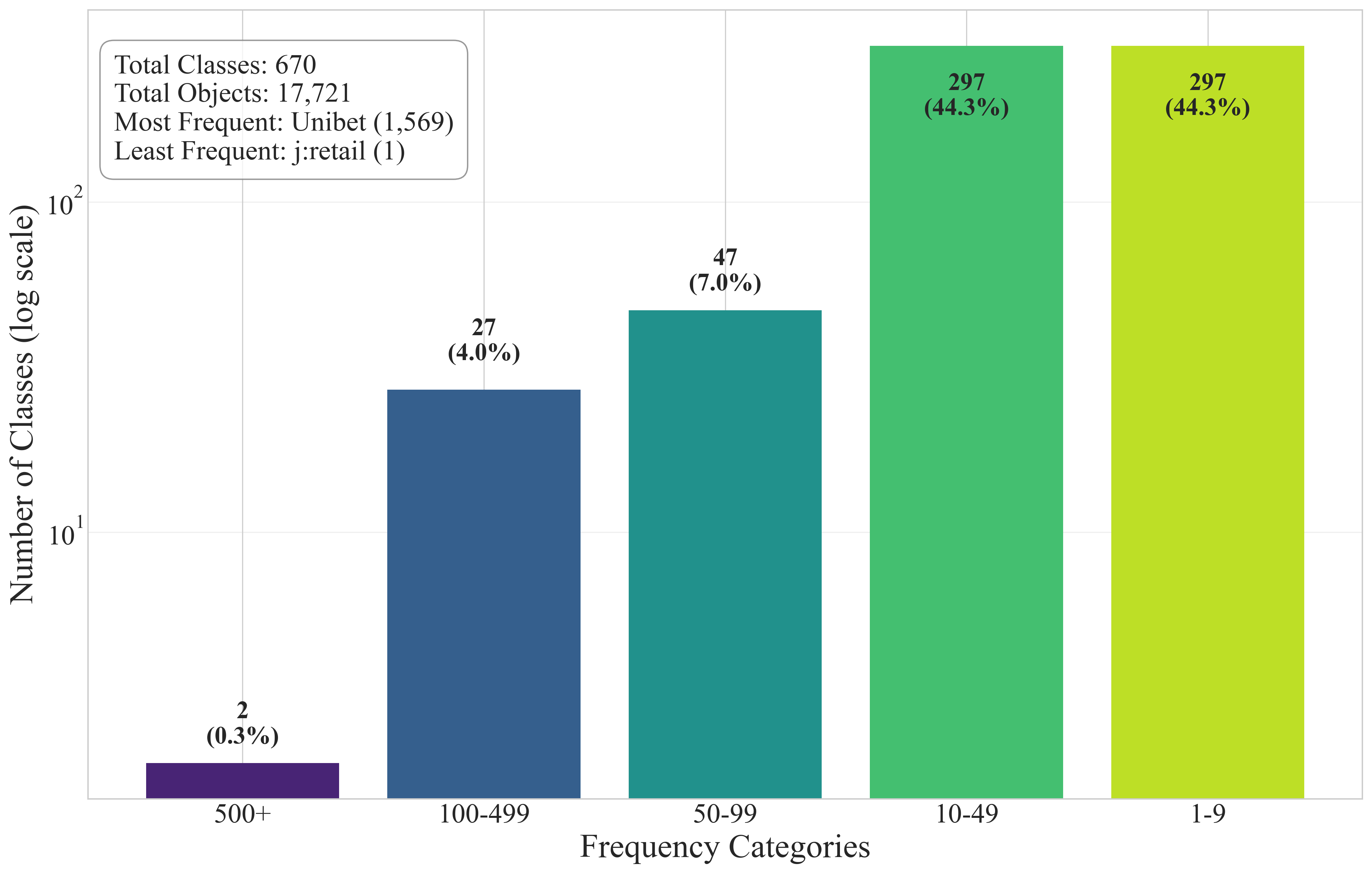}
    \caption{Class frequency distribution: number of classes per instance count category (log scale).}
    \label{fig:class_frequency}
\end{figure}

Logo annotations were performed in Label Studio using \gls{obb}s to capture both geometry and rotation. New classes were added dynamically to account for graphical variants such as standalone icons and text-based wordmarks (e.g., Adidas symbol vs. Adidas text). In total, 670 unique logo classes were annotated, exhibiting a long-tail distribution as shown in Figure~\ref{fig:class_frequency}. This reflects branding disparities in broadcast exposure, with dominant sponsors appearing in over 500 instances. Annotations were converted to YOLO \gls{obb} format, and the dataset was split into training, validation, and test sets (80-10-10). Each split includes paired image and label folders, where label files contain four-corner polygon coordinates per logo instance. To our knowledge, this is the first dataset offering \gls{obb}-based sponsor logo annotations in soccer broadcasts, supporting both detection benchmarks and visibility analytics.

\section{Implementation}\label{sec:implementation}

\begin{figure}[t]
    \centering
    \includegraphics[width=0.9\columnwidth, trim= 0 0 0 0, clip]{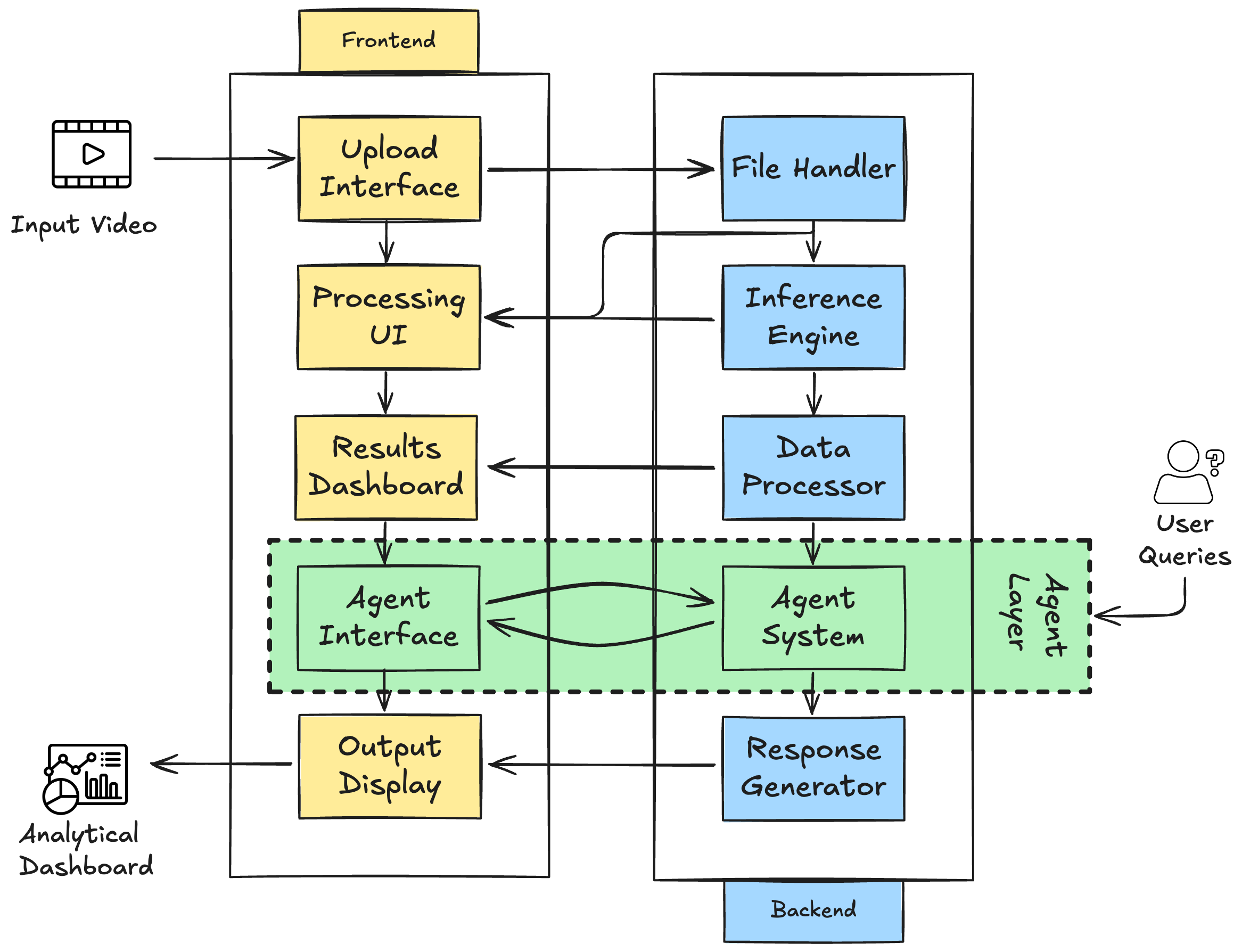}
    \caption {\project modular workflow from input video to analytical dashboard.}
    \label{fig:workflow}
\end{figure}

\begin{figure*}[ht]
    \centering
    \includegraphics[width=0.99\textwidth, trim=100 100 10 50, clip]{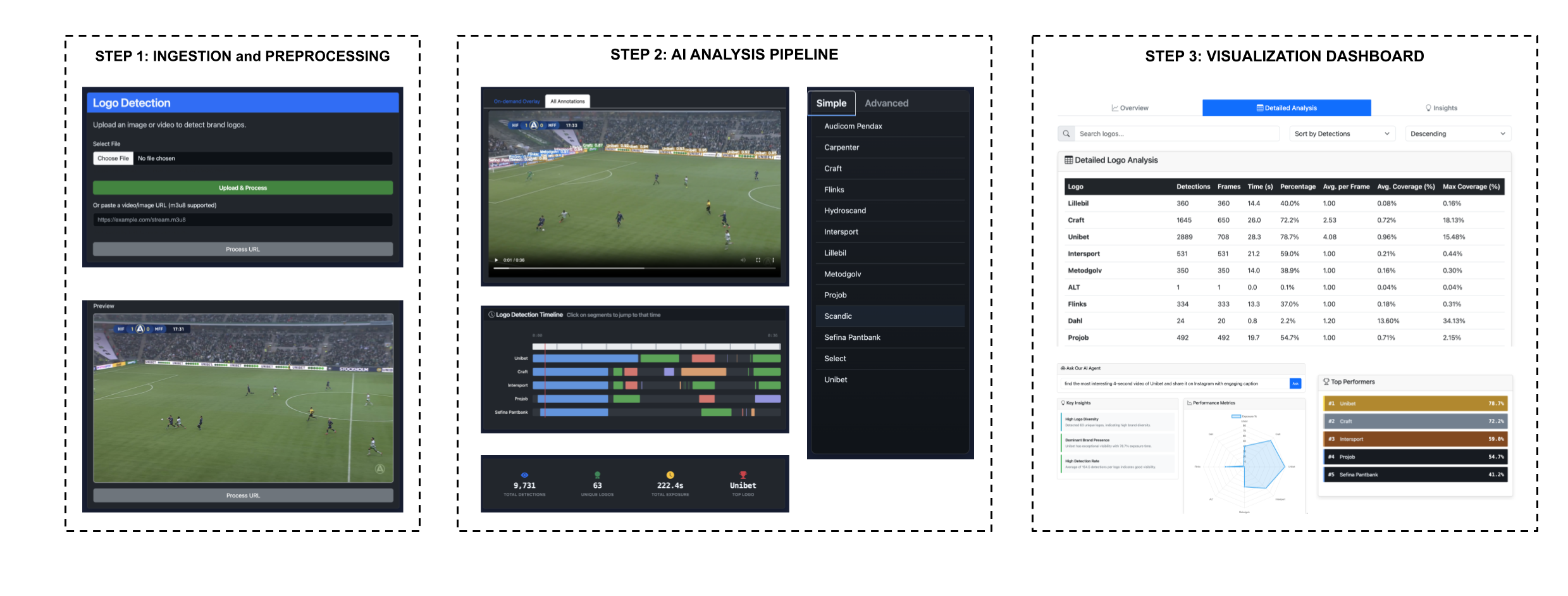}
    \caption{\project dashboard showing video ingestion, logo detection, and sponsorship metric visualization.}
    \label{fig:ui}
    \vspace{-3mm}
\end{figure*}

The proposed \project system transforms raw video into structured analytical insights through a set of interconnected modules organized into functional stages. Components within each stage operate in close coordination to ensure efficient data flow and low-latency performance. Figure~\ref{fig:workflow} illustrates the system architecture, and Figure~\ref{fig:ui} depicts the final user interface of the system.

\subsection{Ingestion and Preprocessing}

The \textbf{Upload Interface} serves as the system’s entry point, allowing users to submit video or image inputs. The interface performs client-side validation before passing files to the \textbf{File Handler}. The file handler executes server-side checks, manages storage, and supports multiple input formats, including MP4, image files, and \gls{hls} playlists. This component standardizes all inputs for subsequent analysis.

\subsection{Automated Analysis Pipeline}

At the core of the backend, the \textbf{Inference Engine} runs the \gls{obb} YOLO model for logo detection. The model processes video frames sequentially, detecting logos under varying scales, rotations, and occlusion conditions, and returns the predicted bounding box coordinates and associated class labels for downstream aggregation and visualization.

The \gls{obb} YOLO model was trained on the dataset described in Section~\ref{sec:dataset}. Six training configurations were evaluated, combining four model sizes (nano: 2.7M parameters, small: 9.7M parameters, medium: 26.4M parameters; large: 44.5M parameters) with two YOLO versions (v8 and v11). All models were trained with an input resolution of $1280\times720$, batch size $32$, and the AdamW optimizer with an initial learning rate of $0.001$. Data augmentation included random rotation, scaling, and contrast adjustment to improve robustness to broadcast variations.

\paragraph{Handling Class Imbalance}

By default, the YOLO classifier uses \gls{bce}~\cite{terven2023loss} with \emph{soft} targets from the task-aligned assigner (i.e., not strictly one-hot) for each anchor--class logit. We also ablated \gls{fl}, which applies a $(1 - p_t)^\gamma$ modulator to down-weight easy examples, where $\gamma>0$ controls suppression strength. In our setup this replaces the default classification term with \gls{fl} on one-hot targets, isolating the effect of \emph{negative suppression} but discarding the assigner-produced \emph{soft} targets that encode match/IoU quality. Thus, while \gls{fl} mitigates negative dominance, it does not exploit quality signals for positives or promote confidence–quality alignment.

We instead adopt \gls{vfl}~\cite{varifocal}, a quality-aware objective for dense, imbalanced detection. \gls{vfl} both (i) down-weights easy negatives to prevent gradient domination and (ii) up-weights positives in proportion to localization quality, aligning class confidence with box accuracy. This focuses learning on informative samples and improves calibration across $670$ long-tailed classes. To formalize this, let $p\in(0,1)$ be the predicted class probability (sigmoid output), $y\in\{0,1\}$ the binary label, and $q\in[0,1]$ a quality target for positives (e.g., derived from the assignment/IoU). The \gls{bce} with a soft target $q$ is:
\[
\mathrm{BCE}(p,q) \;=\; -\,q\log p \;-\; (1-q)\log(1-p).
\]

\gls{vfl} forms a weighted \gls{bce} where negatives and positives are treated differently:

\begin{align*}
\mathcal{L}_{\text{VFL}} = \Big( &\underbrace{\alpha p^{\gamma}(1-y)}_{\substack{\text{down-weights}\\\text{easy negatives}}} + \underbrace{q y}_{\substack{\text{quality-}\\\text{weighted}}} \Big)\times \mathrm{BCE}(p,q)
\end{align*}

with $(\gamma,\alpha)=(2.0,0.75)$. In practice, this means when $y=0$ (negative), the weight $\alpha p^\gamma$ shrinks as $p\!\to\!0$, suppressing the ocean of easy negatives and when $y=1$ (positive), the weight is $q$, so better-localized positives receive stronger gradients, aligning confidence with box quality.

YOLO’s total loss sums over anchors and classes; the classification term simply replaces the original \gls{bce} summand with $\mathcal{L}_{\text{VFL}}$:
\[
\mathcal{L}_{\text{cls}}^{\text{(ours)}} \;=\; \frac{1}{|\mathcal{A}|}\sum_{a\in\mathcal{A}}\sum_{c=1}^{C} \mathcal{L}_{\text{VFL}}\!\big(p_{a,c}, y_{a,c}, q_{a,c}\big),
\]
and the overall OBB objective remains:
\[
\mathcal{L} \;=\; \lambda_{\text{box}}\,\mathcal{L}_{\text{box}}
\;+\; \lambda_{\text{cls}}\,\mathcal{L}_{\text{cls}}
\;+\; \lambda_{\text{dfl}}\,\mathcal{L}_{\text{dfl}}.
\]
$\mathcal{L}_{\text{box}}$ is an oriented-box IoU loss (e.g., rotated/probabilistic IoU), and $\mathcal{L}_{\text{dfl}}$ is the distribution focal loss on discrete distance bins used by modern YOLO heads. Thus, our change is localized to the classification summand. We keep the \gls{obb} regression and distributional terms intact, but swap the classifier’s \gls{bce} for \gls{vfl} to better learn from rare, high-quality positives while damping easy negatives.

Having defined the loss formulation, we next describe the training setup used to evaluate the proposed modification.

\paragraph{Training Setup}
Training was conducted for 200 epochs on three NVIDIA A100 GPUs (80 GB VRAM each) using SLURM-based distributed processing. Among all configurations, YOLOv11-medium achieved the best validation performance and was selected for deployment as the inference engine.

With the inference model established, the \textbf{Data Processor} is introduced to refine raw outputs. It aggregates detections across frames, applies temporal and spatial filtering, and converts results into standardized formats optimized for downstream reasoning, visualization, and storage.

\subsection{Agent-Driven Reasoning}

The \textbf{Agent Interface} (frontend) and \textbf{Agent System} (backend) form an interactive reasoning layer. Since each processed video can yield large volumes of detection data, specialized agents provide task-specific functionality:

\begin{itemize}
    \item \textbf{Analysis Agent}: Interprets all metrics and generates human-readable insights from detection outputs.
    \item \textbf{Highlight Agent}: Identifies and compiles video segments matching user-defined criteria, such as brand-specific exposure above a set threshold.
    \item \textbf{Sharing Agent}: Publishes selected clips directly to social media platforms, handling authentication, authorization, and integration with content-sharing APIs.
    \item \textbf{Coordinator Agent}: Orchestrates multi-step queries by managing agent execution flow. It uses a \gls{llm} with a domain-specific system prompt to convert natural language queries into structured instructions for downstream processing.
\end{itemize}

Agents are implemented using LangGraph, enabling orchestration through defined nodes, edges, and tool integrations. This design supports complex queries, such as: “Find the most exposed 5-second segment for the Unibet logo and post it to Instagram with an engaging caption.” In this case, the coordinator agent triggers the highlight agent to generate the clip and passes the result to the sharing agent for publication.

\subsection{Visualization and Output Delivery}

The \textbf{Processing UI}, \textbf{Results Dashboard}, and \textbf{Output Display} present the analysis. The processing view shows live status and intermediate outputs. The results dashboard provides two synchronized players with a timeline of the top brands by exposure: one renders all \gls{obb} detections, the other highlights a selected brand. The Analytical Dashboard summarizes exposure time, coverage statistics, and detection frequency through compact charts, per-brand rankings, and natural-language queries, providing an interactive information visualization system for sponsor analytics.

\medskip
\noindent\emph{Metric definitions.}
Let the video duration be \(T_v\) seconds, with a frame rate \(r\) (frames per second), yielding \(N = T_v r\) frames in total. Each frame has area \(A_f = W \times H\). For brand \(\ell\) in frame \(i\), we define the visible area \(A_{\ell,i}\) as the sum of all \gls{obb} polygons clipped to the frame.

The relative on-screen coverage is:
\[
c_{\ell,i} = \min\left(1,\; \frac{A_{\ell,i}}{A_f} \right),
\]
and we define an indicator variable:
\[
z_{\ell,i} =
\begin{cases}
1 & \text{if } c_{\ell,i} > 0, \\
0 & \text{otherwise}.
\end{cases}
\]
which equals 1 when the brand is visible in frame \(i\), and 0 otherwise. Using these values, we define the following video-level metrics:

\begin{itemize}
    \item \textbf{Exposure.} The total duration (in seconds) that brand \(\ell\) is visible is given by:
    \[
    E_\ell = \Delta t \sum_{i=1}^{N} z_{\ell,i}, \quad \Delta t = \frac{1}{r},
    \]
    where \(z_{\ell,i}\) indicates brand presence in frame \(i\).

    \item \textbf{Average coverage} is measured in two ways: (i) on frames where the brand appears, and (ii) across all frames regardless of visibility:
    \[
    \bar{C}_\ell^{\text{present}} = 100 \cdot \frac{\sum_i z_{\ell,i} c_{\ell,i}}{\sum_i z_{\ell,i}}, \quad
    \bar{C}_\ell^{\text{overall}} = 100 \cdot \frac{\sum_i z_{\ell,i} c_{\ell,i}}{N}.
    \]

    \item \textbf{Maximum coverage} refers to the highest single-frame percentage occupied by the brand. It is computed by taking the maximum of \(c_{\ell,i}\) over all frames and scaling it to percentage.

    \item \textbf{Detection count} is the total number of OBB detections assigned to the brand across the entire video.
\end{itemize}

\medskip
\noindent
All coverage values are derived from clipped OBB polygons and are capped at \(100\%\) per frame to ensure robustness against overestimation due to overlapping detections.

\section{Results and Evaluation}\label{sec:evaluation}

\subsection{Model Detection Performance}

We evaluated six fine-tuned YOLO-based \gls{obb} configurations, as described in Section~\ref{sec:implementation}, on the held-out test split of our dataset. All models were trained on the proposed dataset using the same training protocol, differing only in backbone size and YOLO version. Performance was assessed using mean Average Precision (mAP) at an IoU threshold of 0.5, together with precision and recall, with results summarized in Table~\ref{tab:obb-results}. Because commercial sponsor-analytics platforms are proprietary and operate on undisclosed datasets and pipelines, direct benchmarking against them is not feasible; we therefore focus on open-method comparisons and methodological contrasts, emphasizing the impact of \gls{obb} versus \gls{hbb} for visibility estimation (see Section~\ref{sec:fit-analysis}).

\begin{figure}[t]
  \centering
  \includegraphics[width=\linewidth]{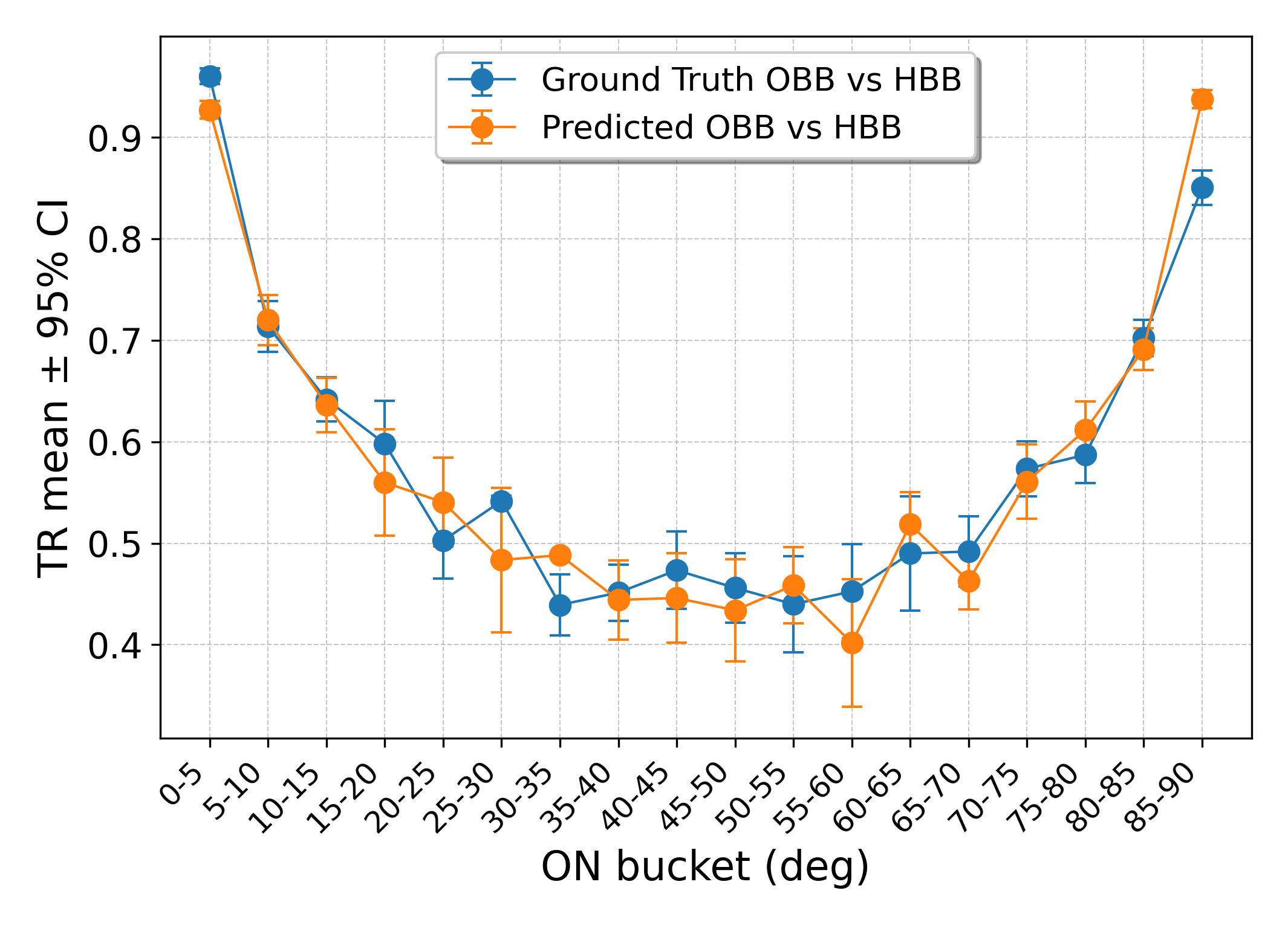}
  \caption{Tightness Ratio vs.\ Orientation Necessity. Higher TR indicates a tighter fit of the OBB relative to its enclosing HBB.}
  \label{fig:tr-vs-on}
\end{figure}

Among all evaluated variants, \textbf{YOLOv11-Medium} achieved the highest mAP@0.5 (\textbf{0.859}) while maintaining high precision (0.96) and competitive recall (0.87). This configuration outperformed the larger YOLOv11 model in average precision despite the latter's slightly higher recall (0.88), suggesting that increased model capacity did not translate into improved precision-recall balance on our dataset. The YOLOv8 models (medium and large) followed closely, with mAP scores of 0.846 and 0.853, respectively, indicating that the shift from YOLOv8 to YOLOv11 yields a measurable but modest improvement when fine-tuned on this domain. In contrast, the YOLOv11-Nano and YOLOv11-Small variants showed reduced mAP (0.781 and 0.817, respectively), reflecting the expected trade-off between computational efficiency and detection robustness.

\begin{figure}[htb]
    \centering
    \includegraphics[width=\linewidth]{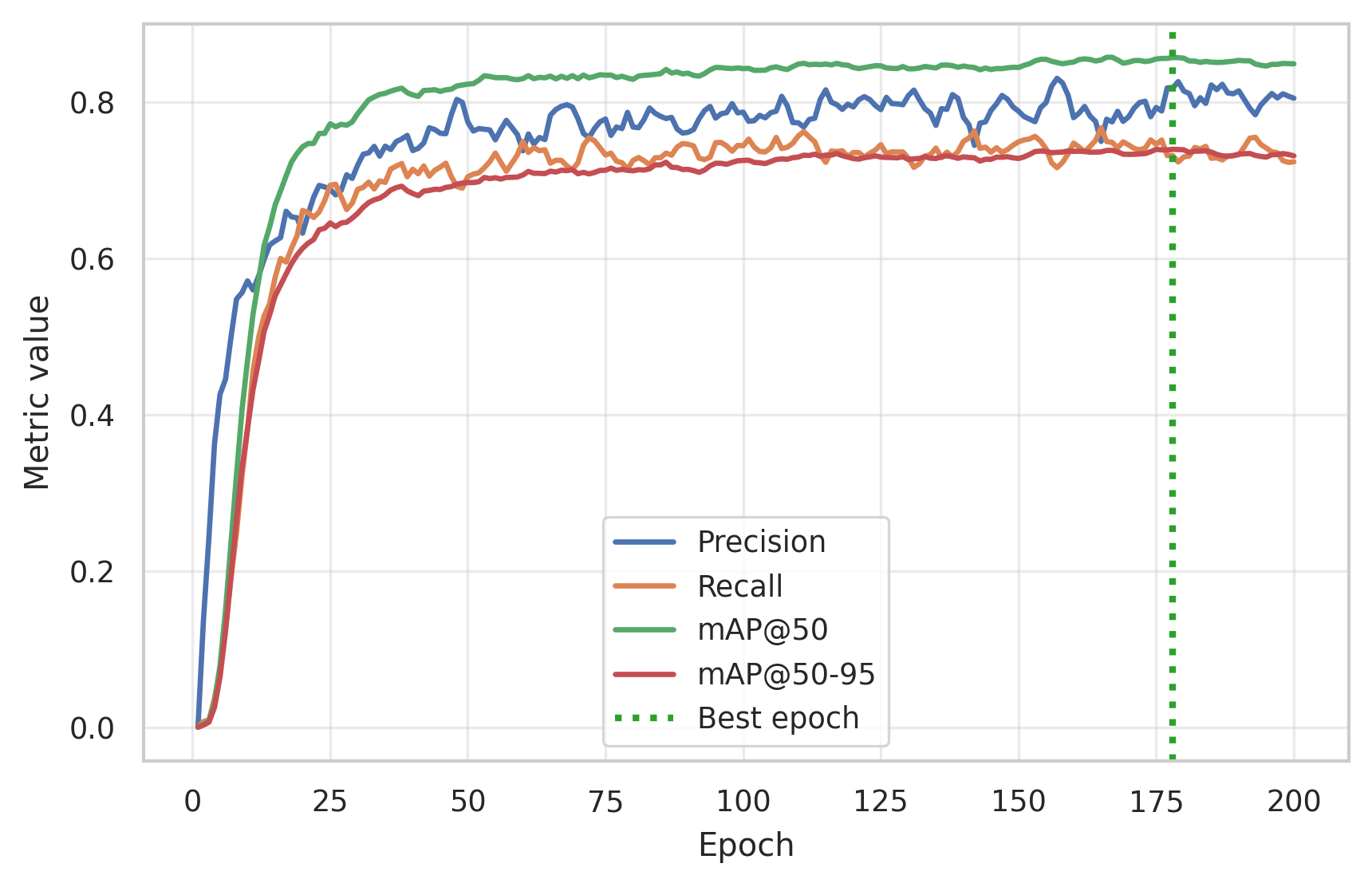}
    \caption{Precision, recall, and mAP@0.5 over training epochs for YOLOv11-medium, showing consistent improvement and convergence.}
    \label{fig:metrics-over-epochs}
\end{figure}

Training dynamics, shown in Figure~\ref{fig:metrics-over-epochs}, reveal a steep rise in mAP@0.5 during the first 20--30 epochs, followed by a gradual plateau toward the optimal epoch (green dashed line). Precision consistently exceeds recall, aligning with the PR curve trend. The mAP@50–95 trajectory stabilizes in the high 0.7 range, confirming robust performance across multiple IoU thresholds, not only at 0.5.

\begin{figure*}[htb]
    \centering
    \begin{subfigure}[t]{0.32\textwidth}
        \centering
        \includegraphics[width=\linewidth]{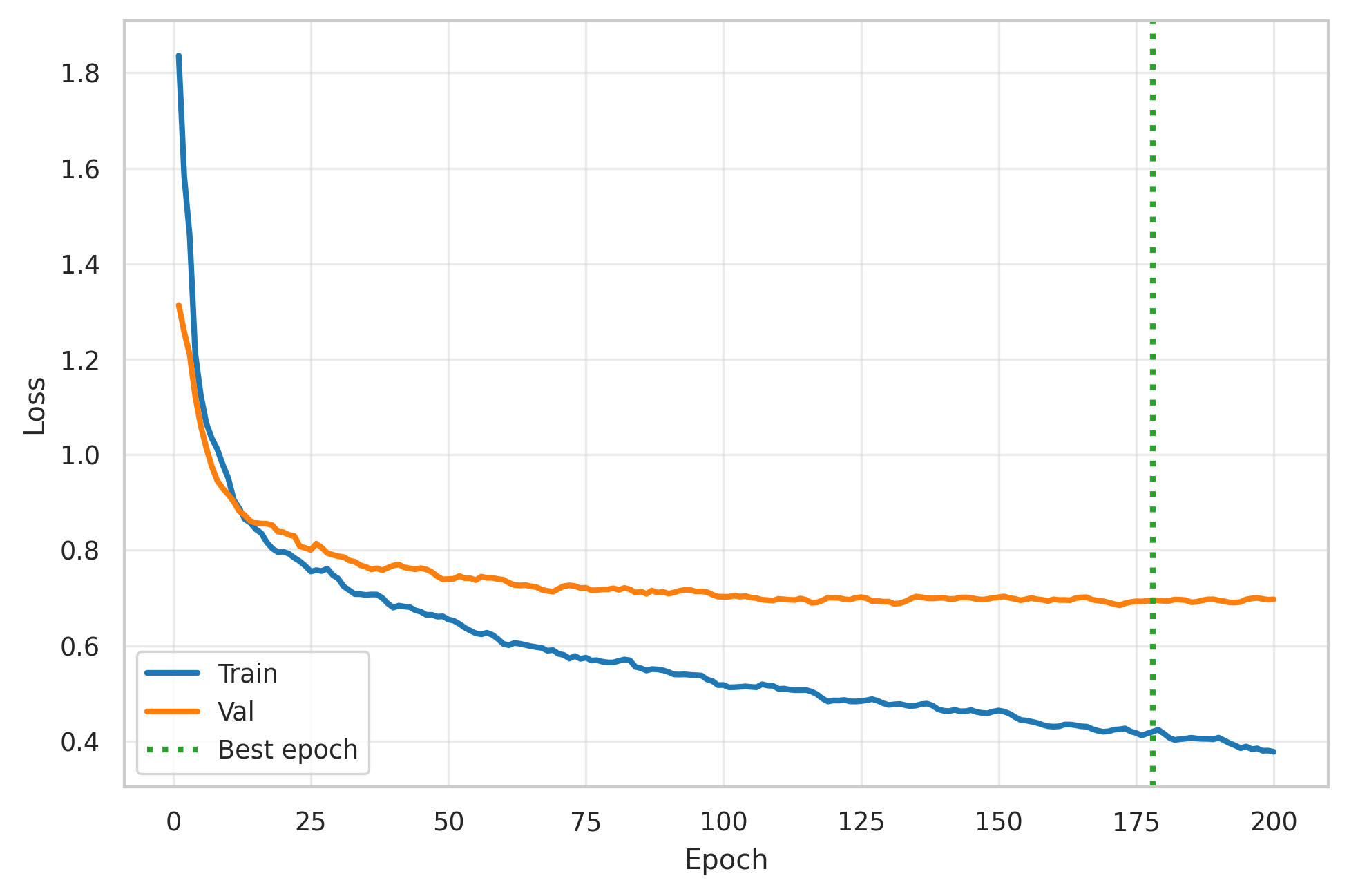}
        \caption{Bounding box regression loss}
        \label{fig:loss-box}
    \end{subfigure}
    \hfill
    \begin{subfigure}[t]{0.32\textwidth}
        \centering
        \includegraphics[width=\linewidth]{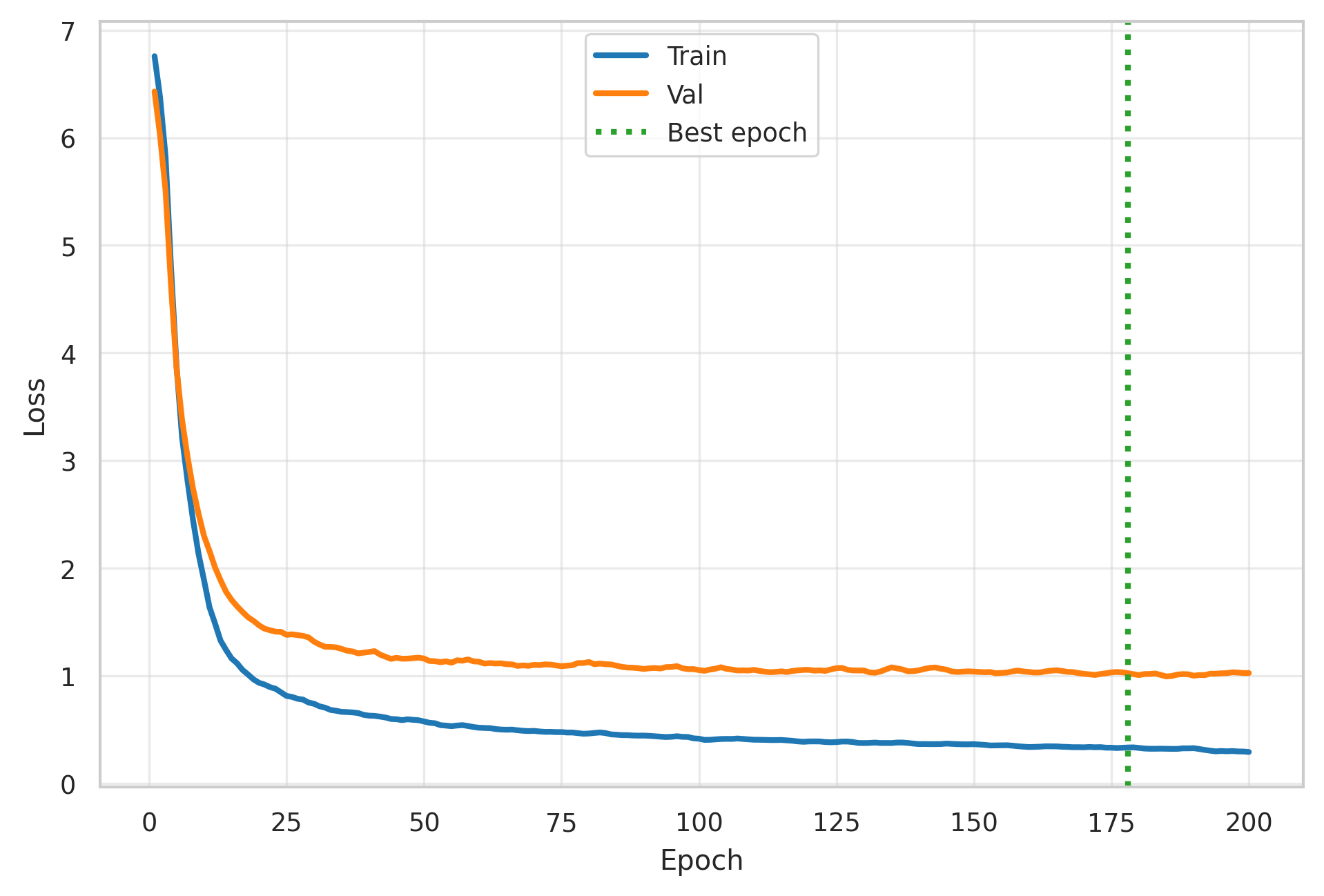}
        \caption{Classification loss}
        \label{fig:loss-cls}
    \end{subfigure}
    \hfill
    \begin{subfigure}[t]{0.32\textwidth}
        \centering
        \includegraphics[width=\linewidth]{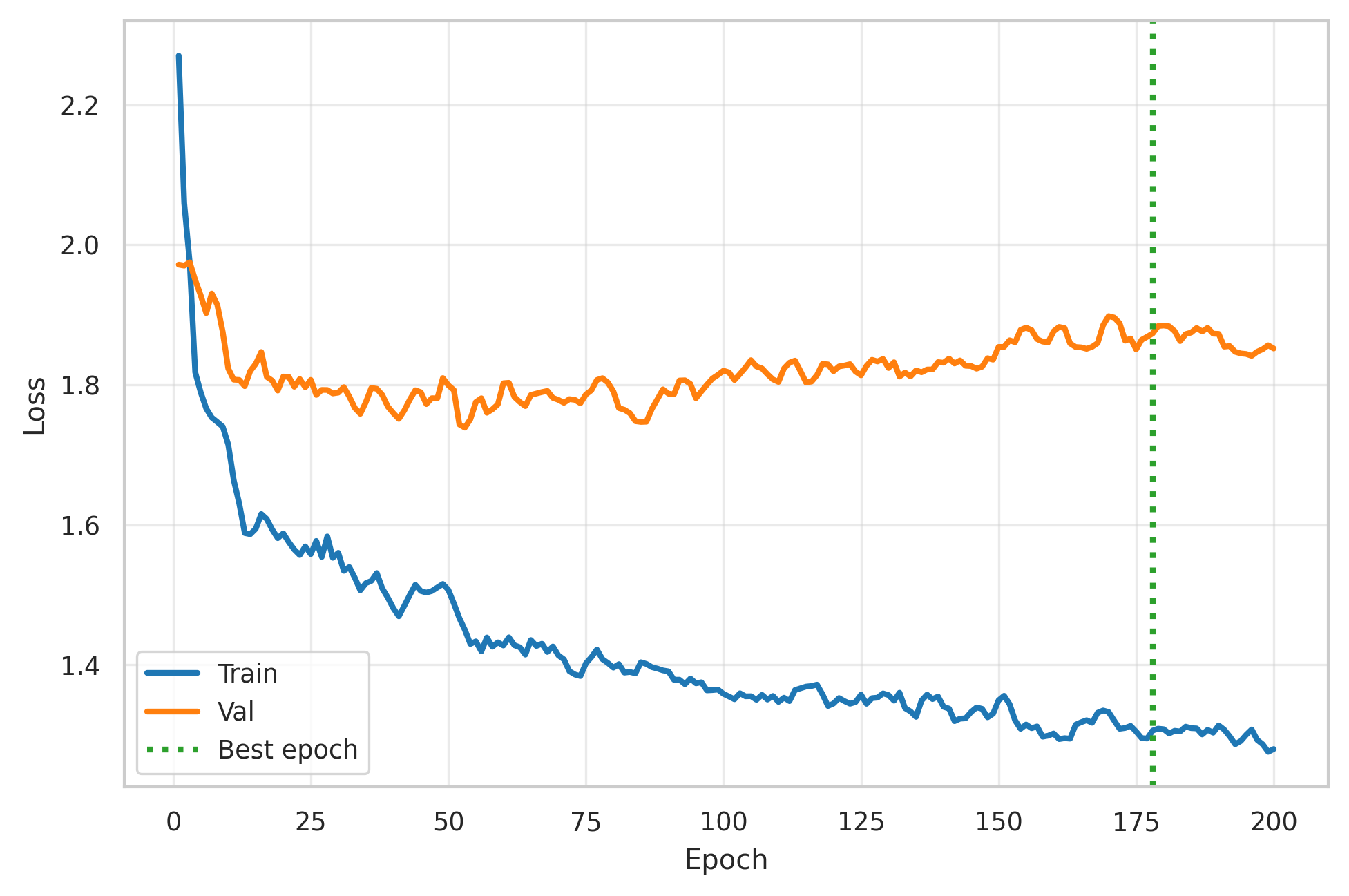}
        \caption{Distribution Focal Loss}
        \label{fig:loss-dfl}
    \end{subfigure}
    \caption{Loss components over training epochs for the YOLOv11-medium model, showing stable convergence for bounding box regression, classification, and distribution focal loss.}
    \label{fig:loss-components}
\end{figure*}

An analysis of the loss components (Figure~\ref{fig:loss-components}) shows that all three bounding box regression, classification, and distribution focal loss drop sharply within the first 20 epochs, indicating rapid acquisition of core detection capabilities. The bounding box regression loss exhibits the smallest gap between training and validation, suggesting robust coarse localization. The classification loss converges earlier, with a modest generalization gap, likely reflecting intra-class variability in logo appearance. The distribution focal loss shows the largest persistent gap, highlighting the difficulty of precise boundary refinement for small, rotated, and partially occluded logos.

Minor oscillations in validation loss during later epochs coincide with slow but steady mAP improvements, indicating ongoing refinement rather than overfitting. This stability across all loss components supports selecting the best epoch as a trade-off between accuracy and generalization.

From a deployment perspective, these findings support a precision-favoured confidence threshold for YOLOv11-Medium, ensuring reliable sponsor visibility measurements while allowing calibrated adjustments toward higher recall if broader coverage is required.

\begin{table}[h]
    \centering
    \caption{Performance comparison of six fine-tuned YOLO-based OBB detection configurations on the test set. Best values in each column are highlighted in bold.}
    \begin{tabular}{lccc}
        \toprule
        Model Variant & mAP@0.5 & Precision & Recall \\
        \midrule
        YOLOv8-Medium   & 0.846   & \textbf{0.96} & 0.88 \\
        YOLOv8-Large    & 0.853   & \textbf{0.96} & 0.88 \\
        YOLOv11-Nano    & 0.781   & 0.95          & 0.85 \\
        YOLOv11-Small   & 0.817   & \textbf{0.96} & 0.86 \\
        \rowcolor{gray!15}
        \textbf{YOLOv11-Medium} & \textbf{0.859} & \textbf{0.96} & 0.87 \\
        YOLOv11-Large   & 0.847   & 0.95          & \textbf{0.88} \\
        \bottomrule
    \end{tabular}
    \label{tab:obb-results}
\end{table}

\subsection{\gls{obb} vs.\ \gls{hbb}: Detection Performance}

To evaluate the effect of bounding box representation on detection performance, we conducted an additional experiment using \gls{hbb} annotations. The dataset, including training, validation, and test splits, was converted into \gls{hbb} format, and the YOLOv11-Medium configuration was retrained using the same hyperparameters as the \gls{obb} setup.

\begin{table}[!t]
    \centering
    \caption{Comparing accuracy of the best-performing \gls{obb} and \gls{hbb} models.}
    \begin{tabular}{lccc}
        \toprule
        Model Type & mAP@0.5 & Precision & Recall \\
        \midrule
        YOLOv11-Medium (\gls{obb}) & 0.859 & 0.96 & 0.87 \\
        YOLOv11-Medium (\gls{hbb}) & 0.865 & 0.95 & 0.88 \\
        \bottomrule
    \end{tabular}
    \label{tab:obb-vs-hbb}
\end{table}

As shown in Table~\ref{tab:obb-vs-hbb}, the \gls{hbb} model achieved a slightly higher mAP@0.5 (+0.6\%), along with marginally higher recall, while the \gls{obb} model yielded a small gain in precision. All differences remain within ±1\%, well inside typical experimental variance. In practical terms, both models perform equally well in detection accuracy and no format demonstrates a meaningful advantage. The choice between them does not impact model performance in any statistically significant way.

Instead, the critical distinction lies in geometric precision. As shown in the \gls{tr} analysis (Section~\ref{sec:fit-analysis}), \gls{obb}s produce tighter enclosures with less background, offering better spatial alignment with actual logo boundaries. This becomes essential for downstream applications such as sponsor visibility measurement, where small differences in area coverage can significantly affect analytics outcomes. In conclusion, while detection metrics are effectively equivalent, the advantage of \gls{obb} lies in its ability to more accurately model the spatial extent of logos.

\subsection{\gls{obb} Alignment with Ground Truth}

To evaluate how well the predicted \gls{obb} align with the annotated ground truth, we compute the rotated Intersection-over-Union (OBB-IoU) between matched predictions and ground-truth boxes in the test set. Unlike axis-aligned IoU, this metric accounts for orientation, providing a more precise measure of geometric agreement.

\begin{figure}[ht]
    \centering
    \includegraphics[width=\linewidth]{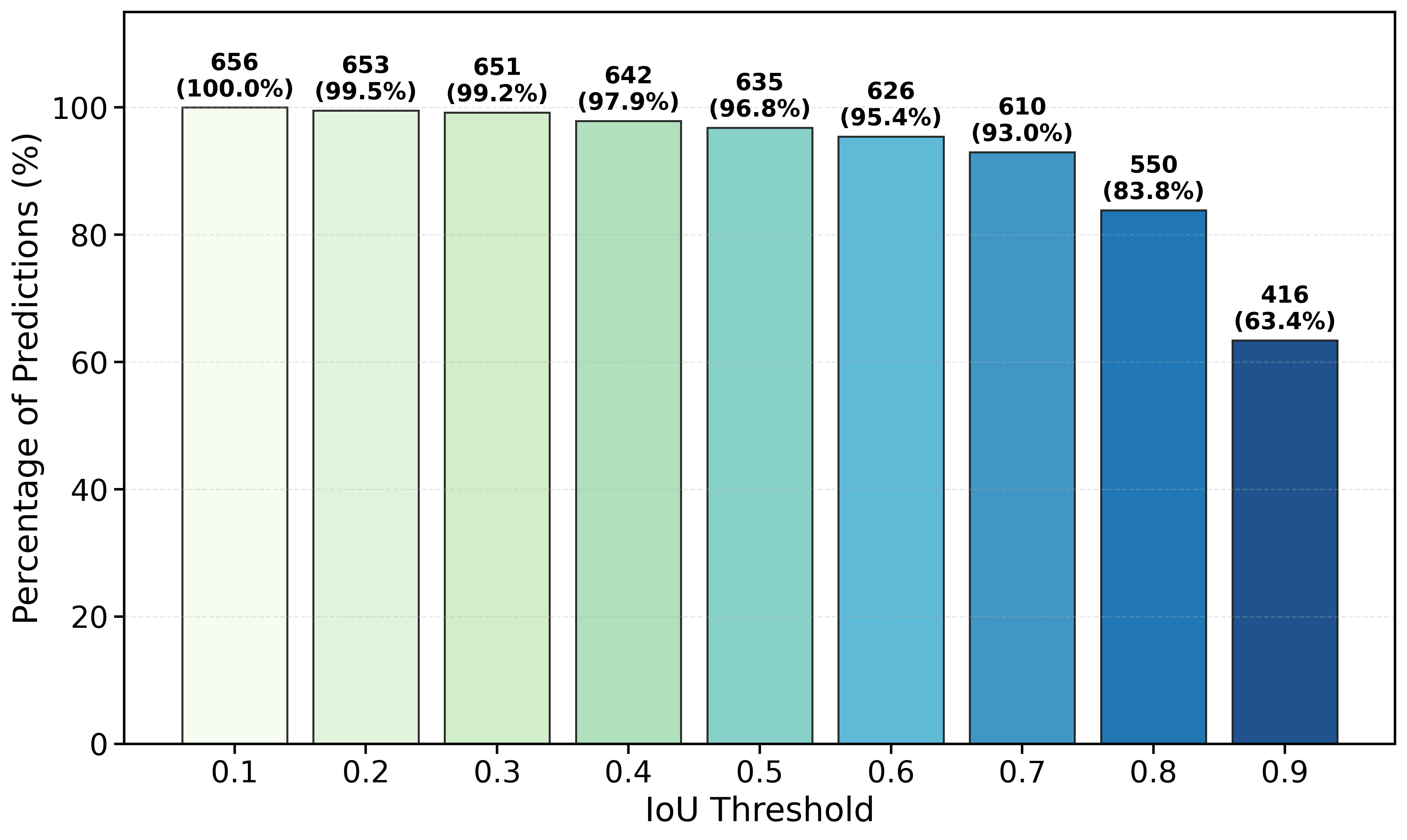}
    \caption{Distribution of predicted–ground-truth OBB-IoU scores. Bars show the percentage of predictions above each threshold, with counts and percentages annotated.}
    \label{fig:iou_analysis}
\end{figure}

Figure~\ref{fig:iou_analysis} shows the percentage of predictions exceeding different IoU thresholds. At the standard detection threshold of 0.5, over 96.8\% of predictions match the ground truth, while 63.4\% achieve very tight alignment with IoU more than 0.9. These results highlight the model's ability to recover both the position and shape of sponsor logos on our test set. This also confirms that the predicted \gls{obb}s closely match the annotated ground truth in classification, position and orientation. The high proportion of predictions above 0.7 IoU (83.8\%) demonstrates the model's robustness in precisely localizing logos across real-world broadcast conditions.

\subsection{OBB vs. HBB: Efficiency Impact of Rotation}\label{sec:fit-analysis}

To assess how effectively \gls{obb}s enclose logo regions while minimizing redundant background compared to \gls{hbb}s, we compute the \gls{tr} using the shoelace formula~\cite{shoelace}, defined as: 

\begin{center}
$\mathrm{TR} = \frac{\mathrm{area(OBB)}}{\mathrm{area(HBB)}}$
\end{center}
Higher \gls{tr} values indicate tighter enclosures with minimal background, while lower values reflect inefficient use of space and excessive inclusion of non-object pixels. Since our dataset contains only oriented annotations and no existing \gls{hbb}-based detector for our logo classes, we derive \gls{hbb}s by computing the minimum enclosing axis-aligned rectangle for each annotated and predicted \gls{obb}. Predicted \gls{obb}s are taken from the model, which achieved the highest mAP in Table~\ref{tab:obb-results}.

We evaluate \gls{tr} for both ground-truth and predicted \gls{obb}s. To analyze the effect of object orientation, we compute the \gls{on} as the absolute angle between the \gls{obb} and the horizontal axis, and group samples into six bins spanning $0^\circ$ to $90^\circ$ in $5^\circ$ increments. As object rotation increases, \gls{hbb}s tend to cover more irrelevant background, leading to lower \gls{tr} scores. Figure~\ref{fig:tr-vs-on} shows the average \gls{tr} per bin with 95\% confidence intervals.

The results show that \gls{tr} is highest for near-horizontal logos, then declines as \gls{on} increases, reaching a minimum near $55^\circ$–$60^\circ$ (\gls{tr} around 0.40), where \gls{hbb}s are least efficient. A slight recovery is observed in the $60^\circ$–$90^\circ$ range due to vertical alignment. The predicted \gls{tr} curve closely follows the ground-truth trend, indicating that the model effectively learns to produce tight, orientation-aware boxes.

\begin{figure}[t]
  \centering
  \begin{subfigure}[t]{0.32\linewidth}
    \centering
    \includegraphics[width=\linewidth, trim=537 0 450 450, clip]{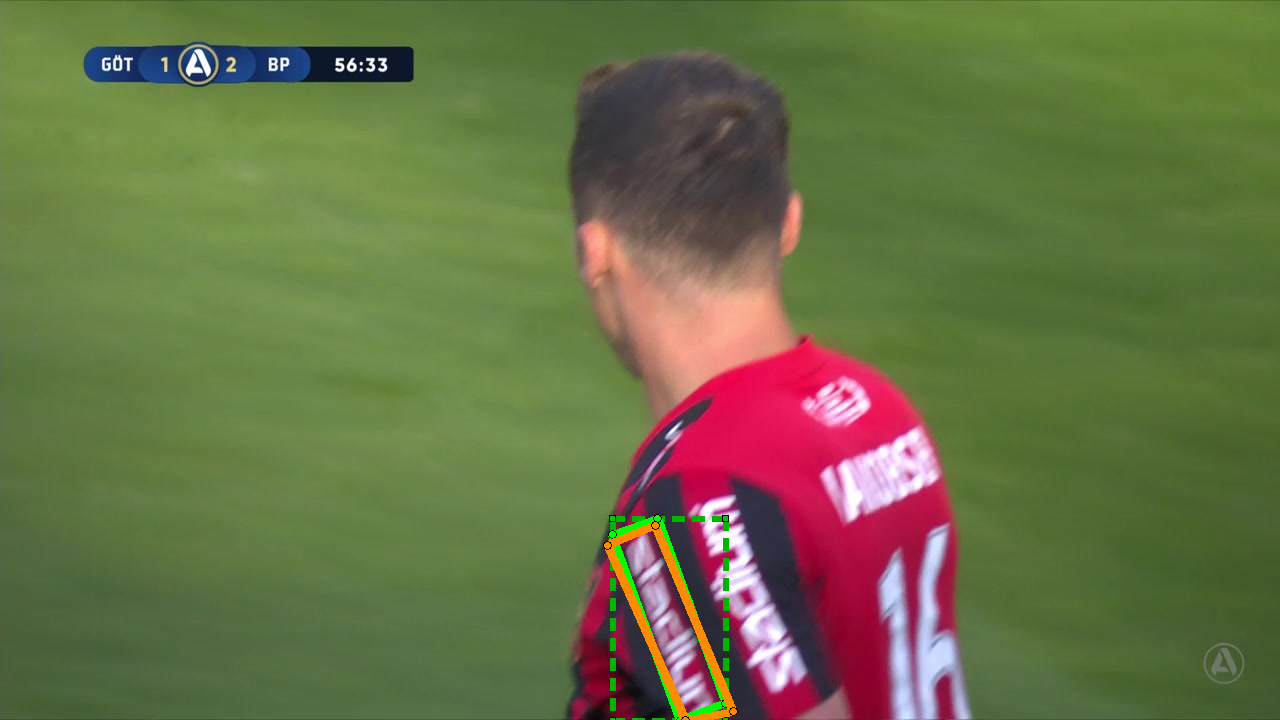}
    \caption{TR=0.41}
  \end{subfigure}
  \hfill
  \begin{subfigure}[t]{0.32\linewidth}
    \centering
    \includegraphics[width=\linewidth, trim=240 0 640 350, clip]{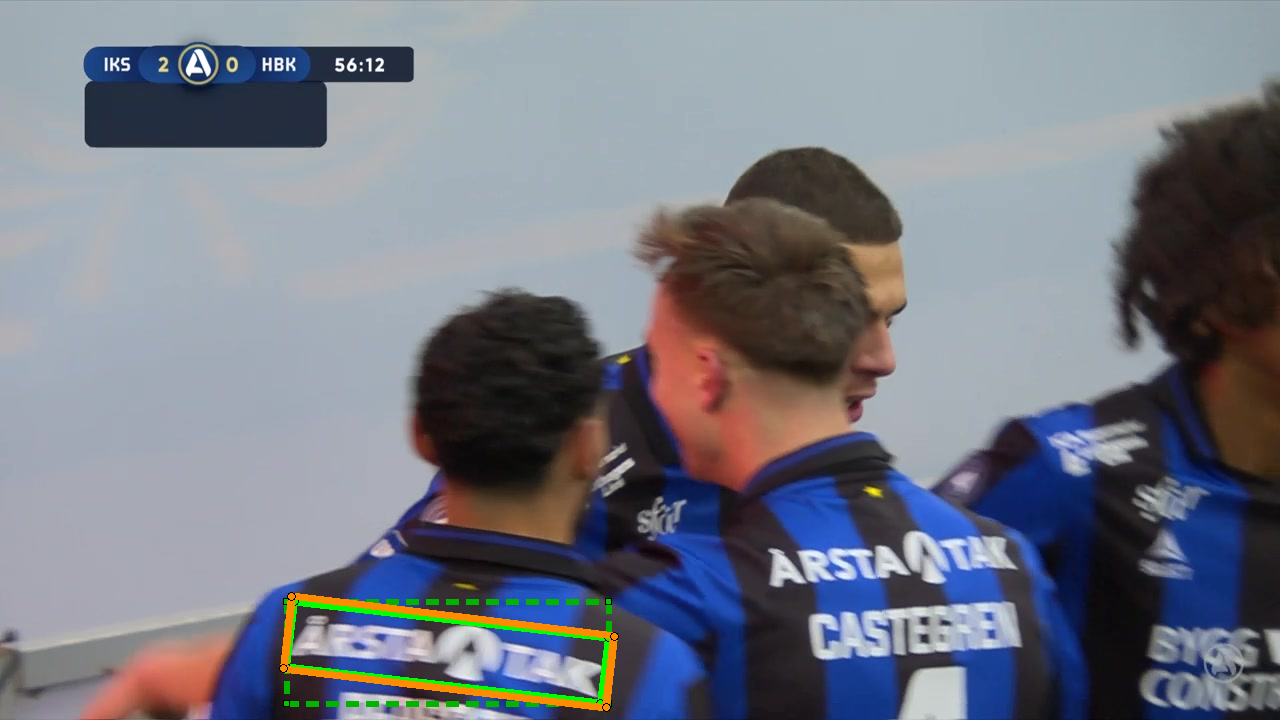}
     \caption{TR=0.63}
   \end{subfigure}
  \hfill
  \begin{subfigure}[t]{0.32\linewidth}
    \centering
    \includegraphics[width=\linewidth, trim=340 180 660 280, clip]{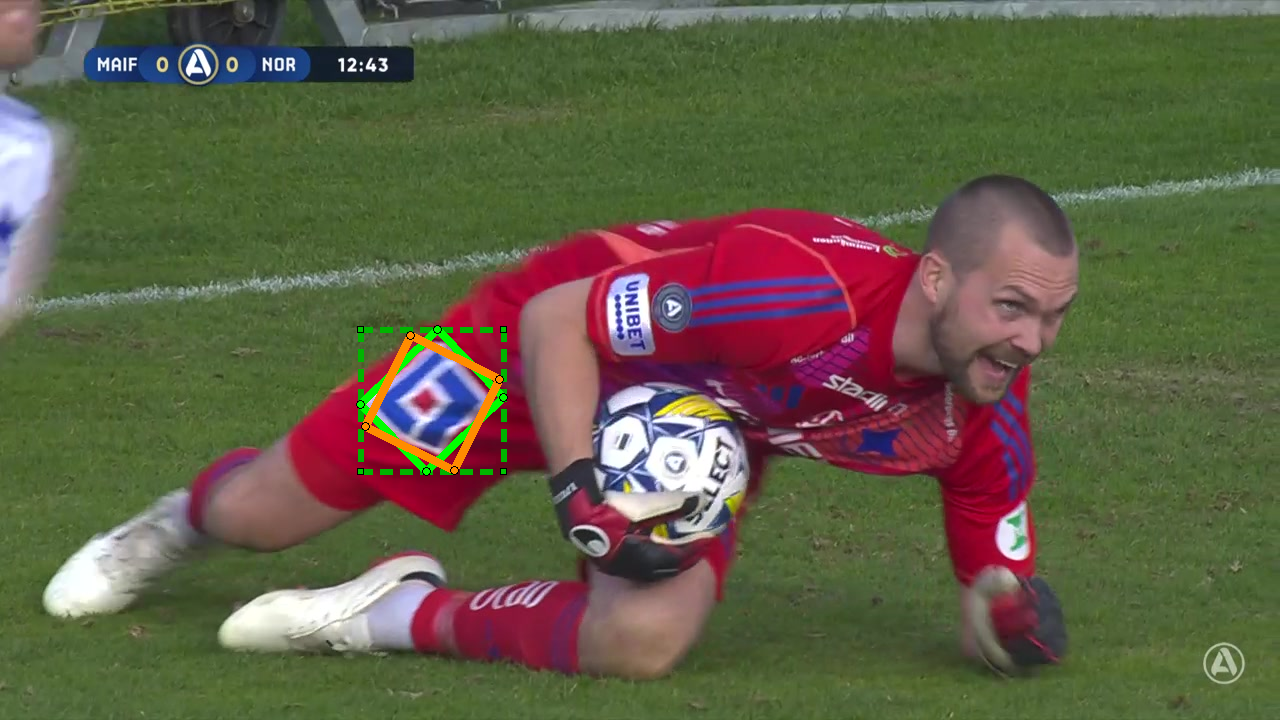}
    \caption{TR=0.49}
  \end{subfigure}
  \caption{
    Example detections with ground-truth OBB (green), enclosing HBB (green dashed), 
    and predicted OBB (orange).
  }
  \label{fig:qual-overlays}
  \vspace{-3mm}
\end{figure}

Qualitative examples in Figure~\ref{fig:qual-overlays} further illustrate how \gls{hbb}s overestimate region size for rotated logos, while \gls{obb}s remain compact and aligned to object geometry.

\subsection{Inference Performance}

Runtime was evaluated on a 30-second broadcast video segment, with each configuration repeated ten times. Table~\ref{tab:inference-performance} summarizes the results across CPU and GPU platforms.
GPU acceleration yields nearly a threefold speedup over CPU-only execution, making the GPU backend more suitable for real-time broadcast applications.

\begin{table}[!b]
    \centering
    \caption{Inference performance across hardware configurations (10-run average).}
    \begin{tabular}{@{}lcccccc@{}}
        \toprule
        Device & Type & 
        \begin{tabular}[c]{@{}c@{}}vCPUs\\Cores\end{tabular} & 
        RAM & VRAM & 
        \begin{tabular}[c]{@{}c@{}}Time\\(ms)\end{tabular} & 
        FPS \\
        \midrule
        MacBook M3    & CPU & 12 & 18~GB & --     & 148.7 & 6.72  \\
        G4dn.xlarge   & GPU & 4  & 16~GB & 16~GB  & 50.0  & 19.98 \\
        \bottomrule
    \end{tabular}
    \label{tab:inference-performance}
\end{table}

\section{Discussion}

This work establishes the value of geometry-faithful localization for sponsor analytics in soccer broadcasts. By predicting \gls{obb}, the system produces detections that align tightly with rotated and foreshortened logos, a common challenge arising from dynamic camera poses, perspective, and fabric deformation. This precision minimizes background leakage, leading to more accurate estimates of on-screen area and position—the foundational inputs for all downstream exposure metrics. The detector achieves strong performance shown in section~\ref{sec:evaluation} that is ideal for business applications where false positives can erroneously inflate sponsor value. Furthermore, with GPU acceleration, the pipeline reaches 19.98 FPS, demonstrating its suitability for near real-time dashboards and automated highlight generation.

Despite strong model performance, the system's accuracy is ultimately bounded by the available data. The Swedish soccer corpus, while realistic, exhibits a long-tail class distribution, which increases variance in per-class AP and limits recall for infrequently seen sponsors. While \gls{vfl} helps by focusing on informative samples, the representational power for rare classes remains a bottleneck. Mitigating this requires a multi-pronged data strategy: expanding coverage across more seasons and leagues, introducing hierarchical labels to group brand variants, employing class-aware sampling and targeted augmentations like copy-paste for rare logos, and leveraging semi-supervised learning on unlabeled broadcasts. Ultimately, robust generalization requires testing across diverse leagues, production styles, and venues, and fostering reproducibility through public configurations and standardized OBB metrics will be key to community adoption and trust.

Beyond single-frame accuracy, robust exposure accounting in video demands temporal coherence. Frame-wise aggregation of detections is susceptible to drift and flicker, where short bursts of false positives or missed detections can introduce noise into final metrics. To address this, \gls{obb}-aware tracking using metrics like polygon-IoU is essential to form stable object tracks. This enables the calculation of track-level exposure, which naturally smooths out transient errors and prevents the overcounting of a single logo appearance. Quantifying the final bias and variance of these aggregated metrics will require rigorous auditing against human-annotated ground truth.

While precise geometric measurement is a critical first step, future work should advance from quantifying presence to assessing value. The impact of a logo's appearance is not uniform; it is highly dependent on context. Exposure during a goal celebration, an event with high replay and sharing potential, is intrinsically more valuable than during routine play. A powerful next step is to develop an event-weighting system that leverages the event metadata already captured during the dataset's creation (e.g., Goal, Shot, Yellow card) to assign higher value to logos appearing in key moments. This can be complemented by a region-of-interest (ROI) analysis. As content is repurposed from 16:9 broadcast aspect ratios to 9:16 vertical formats for social media, logos appearing within the central, action-focused crop hold significantly more value~\cite{smartcrop, sarkhoosh2023soccersocialmedia}. Weighting detections by their proximity to this vertical ROI provides a direct measure of their utility for modern, multi-platform distribution channels, shifting the paradigm from simple measurement to strategic valuation.

The agent layer is the mechanism that translates these advanced valuation models into actionable business intelligence. Its role evolves from a simple content packager to a strategic tool that can automatically identify and surface high-value segments based on the contextual and format-aware metrics described above. For instance, the agent could be tasked to "retrieve all clips of the primary sponsor during goal events that are optimized for vertical sharing." However, this level of automation inherits detector errors and introduces governance challenges. Ensuring reliable and compliant outputs requires defining deterministic tool graphs with comprehensive audit logs, implementing approval gates before external publishing, and evaluating the end-to-end system against task-specific latency budgets and pass/fail criteria. Together, these steps move the system toward higher accuracy, lower latency, and verifiable sponsor analytics at the production scale.

\section{Conclusion}

We introduced \project, a rotation aware system for sponsor visibility in soccer broadcasts that combines oriented object detection with polygon based coverage. The curated dataset with \gls{obb} annotations supports training and evaluation, while a modular processor and a language driven agent layer turn detections into ranked summaries, brand timelines, and shareable clips through synchronized dashboards. The pipeline produces auditable, geometry faithful measurements derived directly from detections and operates at practical throughput with minimal operator effort. Together, the dataset, detector, and agentic analytics provide a coherent foundation for consistent and interpretable sponsor measurement in video.

\section{Acknowledgement}

Funded by the Research Council of Norway, projects \#346671 (AI-Storyteller), \#354154 (Smart Sports Studio), and \#356286 (Forcalytics).

\bibliographystyle{IEEEtran}
\bibliography{references}

\end{document}